\documentclass[11pt,a4paper]{article}
\usepackage[hyperref]{acl2017}
\usepackage{times}
\usepackage{latexsym}

\usepackage{url}

\aclfinalcopy % Uncomment this line for the final submission

\usepackage{color}

\title{Building Chatbots from Forum Data:\\ Model Selection Using Question Answering Metrics}

 \author{Martin Boyanov, Ivan Koychev \\
    Faculty of Mathematics and Informatics\\
    Sofia University ``St. Kliment Ohridski''\\
    Sofia, Bulgaria \\
   {\tt mboyanov@gmail.com}\\
   {\tt koychev@fmi.uni-sofia.bg}\\\And
   Preslav Nakov, Alessandro Moschitti, \\{\bf Giovanni Da San Martino}\\
   Qatar Computing Research Institute\\
   HBKU, Doha, Qatar \\
   {\tt \{pnakov,amoschitti\}@hbku.edu.qa}\\
   {\tt gmartino@hbku.edu.qa}
  }
\date{}

\begin{document}
\maketitle
\begin{abstract}
We propose to use question answering (QA) data from Web forums to train chatbots from scratch, i.e., without dialog training data. First, we extract pairs of question and answer sentences from the typically much longer texts of questions and answers in a forum. We then use these shorter texts to train seq2seq models in a more efficient way. We further improve the parameter optimization using a new model selection strategy based on QA measures. Finally, we propose to use extrinsic evaluation 
%based on mean average precision (MAP) 
with respect to a QA task as an automatic evaluation method for chatbots. 
The evaluation shows that the model achieves a MAP of 63.5\% on the extrinsic task. Moreover, 
%our manual evaluation demonstrates that the model 
it can answer correctly 49.5\% of the questions when they are similar to questions asked in the forum, and 47.3\% of the questions when they are more conversational in style.

%both automatic and manual, show that the model can provide performance up to 63.45\% in MAP and 49.50\% in accuracy in providing correct answers. Additionally, our chatbot tested on completely new and rather different questions shows an accuracy of 47.25\%, according to manual evaluation.
%

%Community Question Answering seeks to automate the process of finding good answers to questions already answered by the community. The standard techniques involve calculating different similarity measures between question and answer and take into account various metadata features about the question and the answer. We propose a way to improve on a state of the art system by using its predictions in a semi-supervised manner to train a seq2seq model able to generate answers given a question. Computing several new features based on the generated responses leads to a 1.35\% increase in F1. 
\end{abstract}

%TODO: take ideas from here: \url{https://docs.google.com/document/d/18SkFsX1KlV6fKdcppnh69xnrG5Rx-JX5YQZC2cK4Osg/edit}

\section{Introduction}
% \begin{itemize}
% \item Intro follow the abstract
% \item description of an informative dialog problem: coherent answers (CA) vs. informative answers (IA).
% \item Problem of S2S: learn coherence but not correctness:
% describe that S2S prefer learning general and less risky answer rather than useful answers.
% \item Our solution based on model selection
% The problem of Community Question Answering(CQA) addresses the problem of finding and ranking comments which answer a given question. We use these question-answer pairs to form a dialog-like corpus. Additionally, we use the state-of-the-art system from the recent competitions to provide relevance judgments for every (question, answer) pair. This allows us to perform model selection based on different ranking metrics which are more aligned with the utility of the seq2seq model. \\
% The results show an indicative improvement on providing \emph{infomative  answers} and, at the same time, it does not lose coherence (according to a small manual evaluation).
% \begin{itemize}
% \item we have a CQA dialog corpus (although not fully dialog style) to learn general QA process.
% \item On this correct and incorrect answers are also available (we have annotated if not blind submission).
% \item we can use this information for model selection using different metrics, cosine, BLEU over them 
% \end{itemize}
% \item the results show highly improvement of the system on providing better answers and, at the same time, it does not loose coherence (according to a small manual evaluation).
% \end{itemize}
	Recently, companies active in diversified business ecosystems have become more and more interested in intelligent methods for interacting with their customers, and even with their employees. Thus, we have seen the development of several general-purpose personal assistants such as Amazon's Alexa,  Apple's Siri, Google's Assistant, and Microsoft's Cortana.
However, being general-purpose, they are not a good fit for every specific need, e.g., an insurance company that wants to interact with its customers would need a new system trained on specific data; thus, there is a need for specialized assistants.

\noindent This aspect is a critical bottleneck as such systems must be engineered from scratch. Very recently, models based on neural networks have been developed, e.g., using seq2seq models \cite{DBLP:journals/corr/VinyalsL15}. Such models provide shallow solutions, but at the same time are easy to train, provided that a large amount of dialog data is available. Unfortunately, the latter is a critical bottleneck as (\emph{i})~the specificity of the domain requires the creation of new data; and (\emph{ii})~this process is rather costly in terms of human effort and time.

%However, we note that large part of the 
Many real-world businesses aiming at acquiring chatbot technology are 
%also typically 
associated with customer services, e.g., helpdesk or forums, where question answering (QA) sections are often provided, sometimes with user evaluation.
Although this data does not follow a dialog format, it is still useful to extract pairs of questions and answers, which are essential to train seq2seq models. Typically, forum or customer care sections contain a lot of content, and thus the requirement about having large datasets is not an issue. The major problem comes from the quality of the text in the pairs that we can extract automatically. One solution is to select data using crowdsourcing, but the task will still be very costly given the required size (hundreds of thousands of pairs)
% for training good models) 
and its complexity.

In this paper, we propose to use data extracted from a standard question answering forum %Qatar Living, \footnote{url{http://www.qatarliving.com}} 
for training chatbots from scratch. The main problem in using such data is that the questions and their associated forum answers are noisy, i.e., not all answers are good. Moreover, many questions and answers are very long, e.g., can span several paragraphs. This prevents training effective seq2seq models, which can only manage (i.e., achieve effective decoding for) short pieces of text.

\noindent We tackle these problems by selecting a pair of sentences from each questions--answer pair, using dot product over averaged word embedding representations.
%of both the questions and the answers. 
The similarity works both (\emph{i})~as a filter of noisy text as the probability that random noise occurs in the same manner in both the question and the answer is very low, and (\emph{ii})~as a selector of the most salient part of the user communication through the QA interaction.

We further design several approaches to model selection and to the evaluation of the output of the seq2seq models. The main idea is, given a question, (\emph{i})~to build a classical vector representation of the utterance answered by the model, and (\emph{ii})~to evaluate it by ranking the answers to the question provided by the forum users. We rank them using several metrics, e.g.,~the dot product between the utterance and a target answer. This way, we can use the small training, development and test data from a SemEval task \cite{nakov-EtAl:2016:SemEval2} to indirectly evaluate the quality of the utterance in terms of Mean Averaged Precision (MAP). Moreover, we use this evaluation in order to select the best model on the development set, while training seq2seq models.

We evaluate our approach using (\emph{i})~our new MAP-based extrinsic automatic evaluation on the SemEval test data, and (\emph{ii})~manual evaluation carried out by four different annotators on two sets of questions: from the forum and completely new ones, which are more conversational but still related to the topics discussed in the forum (life in Qatar). The results of our experiments demonstrate that our models can learn well from forum data, achieving MAP of 63.45\% on the SemEval task, and accuracy of 49.50\% on manual evaluation. Moreover, the accuracy on new, conversational questions drops very little, to 47.25\%, according to our manual evaluation.

% \alex{\sout{
% The remainder of this paper is organized as follows. 
% Section~\ref{sec:relatedwork} discusses related work. 
% Section~\ref{sec:method} describes the target forum data and how we converted it for training a seq2seq model. Section~\ref{sec:model} presents (\emph{i})~our new model selection approach and (\emph{ii})~our new evaluation method.
% Section~\ref{sec:experiments} reports on the results of our experiments using the automatic evaluation of our model selection approaches.
% Section~\ref{sec:erroranalysis} illustrates a manual evaluation of our systems analyzing some of the answers the system provides.
% Finally, Section~\ref{sec:conclusion} draws some conclusions. 
% }}
\section{Related Work} \label{sec:relatedwork}
% \alex{Giovanni/Preslav/Alex}
% \begin{itemize}
% \item Dialog systems, seq2seq
% \item CQA
% \end{itemize}

Nowadays, there are two main types of dialog systems: sequence-to-sequence and retrieval-based. Here we focus on the former. Seq2seq is a kind of neural network architecture, initially proposed for machine translation \cite{sutskever2014sequence,cho-EtAl:2014:EMNLP2014}. Since then, it has been applied to other tasks such as text summarization \cite{summarization}, image captioning \cite{Vinyals:2017:STL:3069214.3069249}, and, of course, dialog modeling \cite{shang-lu-li:2015:ACL-IJCNLP,li-EtAl:2016:N16-11,gu-EtAl:2016:P16-1}.
%DBLP:journals/corr/VinyalsTBE14,

%The seq2seq architecture has two primary components: an encoder and a decoder. The encoder consists of a recurrent neural network (RNN), which processes the input sequence one token at a time and updates its hidden layer. The assumption is that the final state of the hidden layer encodes the semantics of the input sequence and is sufficient to produce a result in the domain of the target sequence. 
\noindent The initial seq2seq model assumed that the semantics of the input sequence can be encoded in a single vector, which is hard, especially for longer inputs. Thus, attention mechanisms have been introduced \cite{DBLP:journals/corr/BahdanauCB14}. This is what we use here as well.

Training seq2seq models for dialog requires large conversational corpora such as Ubuntu  \cite{DBLP:conf/sigdial/LowePSP15}. Unstructured conversations, e.g., from Twitter, have been used as well \cite{sordoni-EtAl:2015:NAACL-HLT}. See \cite{DBLP:journals/corr/SerbanLCP15} for a survey of corpora for dialog. Unlike typical dialog data, here we extract, filter, and use question-answer pairs from a Web forum.

An important issue with the general seq2seq model is that it tends to generate general answers like \emph{I don't know}, which can be given to many questions. This has triggered researchers to explore diversity promotion objectives \cite{li-EtAl:2016:N16-11}. Here, we propose a different idea: select training data based on performance with respect to question answering, and also optimize with respect to a question answering task, where giving general answers would be penalized.

It is not clear how dialog systems should be evaluated automatically, but it is common practice to use BLEU \cite{Papineni:2002:BMA:1073083.1073135}, and sometimes Meteor \cite{Lavie:2007:MAM:1626355.1626389}: after all, seq2seq models have been proposed for machine translation (MT), so it is natural to try MT evaluation metrics for seq2seq-based dialog systems as well. However, it has been shown that BLEU, as well as some other popular ways to evaluate a dialog system, do not correlate well with human judgments \cite{liu-EtAl:2016:EMNLP20163}. Therefore, here we propose to do model selection as well as evaluation extrinsically, with respect to a related task: Community Question Answering.

\section{Data Creation}
\label{sec:method}
%\alex{Giovanni}

In order to train our chatbot system, we converted an entire Community Question Answering forum into a set of question--answer pairs, containing only one selected sentence for each question and for each answer.\footnote{We released the data here: \url{http://goo.gl/e6UWV6}} We then used these selected pairs in order to train our seq2seq models. 
Below, we describe in detail our data selection method along with our approach to question-answer sentence pair selection.

\subsection{Forum Data Description}
We used data from a SemEval task on Community Question Answering \cite{nakov-EtAl:2015:SemEval,nakov-EtAl:2016:SemEval2,SemEval-2017:task3}. 
The data consists of questions from the Qatar Living forum\footnote{Qatar Living: \url{http://www.qatarliving.com}} and a (potentially truncated) thread of answers for each question. Each answer is annotated as \textit{Good}, \textit{Potentially Useful} or \textit{Bad}, depending on whether it answers the question well, does not answer well but gives some potentially useful information, or does not address the question at all (e.g., talks about something unrelated, asks a new question, is part of a conversation between the forum users, etc.). The goal of the task is to rank the answers so that \textit{Good} answers are ranked higher than \textit{Potentially Useful} and \textit{Bad} ones. 
The participating systems are evaluated using 
Mean Average Precision (MAP) as the official evaluation metric. 

The data for SemEval-2016 Task 3, subtask A comes split into training, development and test parts with 2,669/17,900, 500/2,440 and 700/3,270 questions/answers, respectively. In addition, the task organizers provided raw unannotated data, which contains 200K questions and 2M answers. Thus, our QA data consists of roughly 2M answers extracted from the forum. We paired each of these answers with the corresponding question in order to make training question--answer pairs for our seq2seq system.  We made sure that the development and the testing datasets for SemEval-2016 Task 3 were excluded from this set of training question--answer pairs.

We used the annotated development and test sets both to carry out 
our new model selection, as explained in Section~\ref{modsel},
and our new evaluation, as described in Section~\ref{sec:experiments}. Both model selection and our proposed evaluation are based on extrinsic evaluation with respect to the SemEval task.

%a classifier to predict whether an answer is \textit{Good} vs. not-Good (i.e., \emph{Bad} or \emph{Potentially Useful}). We then applied this classifier to the entire raw dataset. Finally, we used the \emph{Good} question-answer pairs to train our seq2seq model.

%Our Good vs.~not-Good classifier\footnote{We omit the details about this classifier to save space, as it is not essential for the method we present here.} has about 80\% accuracy, which is on par with the best systems at SemEval-2016 Task 3.

% \subsubsection{Forum Dialog}
% \begin{itemize}
% \item Description of Qatar Living data
% \item a figure (tab) with an example of dialog
% \end{itemize}

%\subsubsection{Measuring QA and Dialog systems}
%evaluation measures: BLEU score, MAP?, accuracy

%The main focus was on the unannotated QatarLiving data provided for SemEval Task 3. The best system from 2016 was used to provide scores for every (question, answer) pair in the dataset. Afterwards, a sentence selection scheme was used to specify (question, answer) pairs on a sentence level.

\subsection{Sentence Pair Selection}
As the questions and the comments\footnote{In our forum, the comments are considered as answers.} in Qatar Living can be quite long, we reduced the question-answer pairs to single-sentence pairs. In particular, given a question-answer pair, we first split the question and the answer from the pair into individual sentences, and then we computed the similarity between each sentence from the question and each sentence from the answer. Ultimately, we kept the most similar pair. %Note that we can select different sentences from the question when comparing it to different \emph{Good} answers.

\noindent We measured the similarity between two sentences based on the cosine between their embeddings. We computed the latter as the average of the embeddings of the words in a sentence. We used pre-trained word2vec embeddings \cite{Mikolov:2013:DRW:2999792.2999959,mikolov-yih-zweig:2013:NAACL-HLT} 
fine-tuned\footnote{\url{https://github.com/tbmihailov/semeval2016-task3-CQA\#resources}} for Qatar Living \cite{mihaylov-nakov:2016:SemEval}, and proved useful in a number of experments with this dataset \cite{guzman-marquez-nakov:2016:P16-2,hoque-EtAl:2016:COLINGDEMO,Mihaylov:2017:LGP:3077136.3080757,mihaylova-EtAl:2016:SemEval,nakov-marquez-guzman:2016:EMNLP2016}.

More specifically, we generated the vector representation for each sentence by averaging 300-dimensional word2vec vector representations after stopword removal. 
We assigned a weight to the word2vec vectors with TF$\times$IDF, where IDF is derived from the entire dataset. Note that averaging has the consequence of ignoring the word order in the sentence. We leave for future work the exploration of more sophisticated sentence representation models, e.g., based on long short-term memory \cite{Hochreiter:1997:LSTM} and convolutional neural networks \cite{kim:2014:EMNLP2014}.

% \alex{
% \subsection{Pre-training on Ubuntu Dialog Data}
% \sout{
% Initial experiments have shown that training only on the (question, answer) pairs does not lead to a good general conversational experience. Chatbots trained on this data, would fail some simple handshake steps like saying \emph{hello} or \emph{no problem}.  In order to improve the general language capabilities, we experimented with pre-training our seq2seq model on v.2 of the Ubuntu Dialog Corpus }\footnote{\url{https://github.com/rkadlec/ubuntu-ranking-dataset-creator}}   \cite{DBLP:conf/sigdial/LowePSP15}\sout{, and then we continued training on our data. 
% Manual evaluation has shown that this yields improvements in the general conversational capabilities of our chatbot. }
% %, but did not improve our main task to provide \good  answers. Thus, all results below skip the Ubuntu pretraining step.
% 
% }

\section{Model Selection and Evaluation}
\label{sec:model}

In this section, we describe our approach to automatic evaluation as well as model selection for seq2seq models.

\subsection{Evaluation}

\paragraph{Intrinsic evaluation.} We evaluated our model \emph{intrinsically} using BLEU as is traditionally done in dialog systems.

\paragraph{Extrinsic evaluation.} We further performed \emph{extrinsic} evaluation in terms of how much the answers we generate can help solve the SemEval CQA task. In particular, we input each of the test questions from SemEval to the trained seq2seq model, and we obtained the generated answer. Then, we calculated the similarity, e.g., TF$\times$IDF-based cosine (see below for more detail), between that seq2seq-generated answer and each of the answers in the thread, and we ranked the answers in the thread based on this similarity. Finally, we calculated MAP for the resulting ranking, which evaluates how well we do at raning the \textit{Good} answers higher than the not-Good ones (i.e., \emph{Bad} or \emph{Potentially Useful}). 
As a baseline, we used the MAP ranking produced by comparing the answers to the question (instead of the generated answer). 
%\alex{We also experimented with a variant of this extrinsic evaluation, where we sum the different similarity scores with the qc-sim similarity scores and we ranked based on this sum.}

\subsection{Model Selection} \label{sec:seq2seq}
\label{modsel}

\label{sec:modelselection}
The training step produces a model that evolves over the training iterations. We evaluated that model after each 2,000 minibatch iterations. Then, among these evaluated models, we selected the best one, which we used for the test set. 
We used three model selection approaches, optimizing for MAP and for BLEU calculated on the development set of the SemEval-2016 Task 3, and for the seq2seq loss on the training dataset.

\begin{table*}[ht]
  \centering
  \begin{tabular}{|c|l|cc|cc|}
      \hline
%   \multicolumn{6}{|c|}{\bf Larger seq2seq Model}\\\hline % Combined-recalculated-3
  & \bf Optimizing for & \bf MAP  & \bf BLEU  & \bf Iteration & \bf Ans. Len. \\
  \hline  
1 & MAP       & 63.45 & 9.18    & 192,000 &   10.56 \\
2 & BLEU      & 62.64 & 8.16    & 16,000  &   16.31 \\
3 & seq2seq loss  & 62.81 & 7.00    & 200,000 & 8.73  \\ \hline\hline
4 & Baseline    & 52.80 & -     & -     & - \\
\hline
\end{tabular}
  \caption{Evaluation results using the seq2seq model and optimizing during training for MAP (on DEV) vs. BLEU (on DEV) vs. the seq2seq loss (on TRAIN). The following columns show some results on TEST when selecting the best training model on DEV (for MAP and BLEU) and on TRAIN (for the seq2seq loss). We report BLEU and MAP, as well as the iteration at which the best value was achieved on DEV/TRAIN, and the average length of the generated answers on TEST. \label{tab:results}}
\end{table*}

\paragraph{Seq2seq loss.}
We consider the loss that the seq2seq model optimizes during the training phase. Notice that in this case no development set is required for model selection. 

\paragraph{Machine translation evaluation measure (BLEU).}
A standard model selection technique for seq2seq models is to optimize BLEU. Here, we calculated multi-reference BLEU between the generated response and the \emph{Good} answers in the thread (on average, there are four \emph{Good} answers out of ten in a thread). We then take the average score over all threads in the development set. 

\paragraph{Extrinsic evaluation based on MAP.}
The main idea for this model selection method is the following: given a question, the seq2seq model produces an answer, which we compare to each of the answers in the thread, e.g., using cosine similarity (see below for detail), and we use the score as an estimation of the likelihood that an answer in the thread would be good. In this way, the list of candidate comments for each question can be ranked and evaluated using MAP. 
We used the gold relevancy labels available in the development dataset to compute MAP.
 
More formally, given an utterance $u_q$ returned by the seq2seq model in response to a forum question, we rank the comments $c_1,\ldots,c_{n}$ from the thread according to the values $r(u_q, c_i)$. 
We considered the following options for $r(u_q,c_i)$: 

\begin{itemize}
\item[-] \textbf{cos}: this is the cosine between the embedding vectors of $u_q$ and $c_i$, where the embeddings are calculated as the average of the embedding vectors of the words, using the fine-tuned embedding vectors from \cite{mihaylov-nakov:2016:SemEval}; 
\item[-]  \textbf{BLEU}: this is the sentence-level BLEU+1 score between $u_q$ and $c_i$; 
\item[-] \textbf{bm25}: this is the BM25 score ~\cite{Robertson:2009:PRF:1704809.1704810} between $u_q$ and $c_i$;
\item[-] \textbf{TF$\times$IDF}: we build a TF$\times$IDF vector, where the TF is based on the frequency of the words in $u_q$, and the IDF is calculated based on the full SemEval data (all 200K questions and all 2M answers), we then repeat the procedure to obtain a vector for $c_i$, and finally we compute the cosine between these two vectors; 
\end{itemize}

%We define the \textbf{baseline} as the ranking produced by using the tfidf-cosine similarity between question and comment. \\
%We also define a variant of each $r()$ defined above in which the similarity score of the metric is summed with the baseline similarity score and then the responses are ranked based on these scores (\textbf{+baseline}). 

We also define a variant of each of the $r(x,y)$ functions above, where the similarity score is further summed with the TF$\times$IDF-cosine similarity between the question and the comment (\textbf{+qc-sim}). 
Finally, we define yet another metric, \textbf{Avg}, as the average of all $r()$ functions defined in this section. 

\section{Experiments}\label{sec:experiments}

We compare the model selection approaches described in Section~\ref{sec:modelselection} above, with the goal to devise a seq2seq system that gives fluent, \emph{good} and informative answers, i.e., avoids answers such as \textit{``I don't know''}. 
%Thus, we investigate how the training procedure could be improved towards this goal. 

%Our experiments investigate how the seq2seq training procedure could be improved, so that its utterances are not only syntactically correct and fluent, but also \good answers the question. 
%Specifically, we compare the model selection approaches described in Section~\ref{sec:modelselection} above. 

\subsection{Setup}
Our model is based on the seq2seq implementation in TensorFlow. However, we differ from the standard setup in terms of preprocessing, post-processing, model selection, and evaluation.

First, we learned subword units using byte pair encoding \cite{DBLP:journals/corr/SennrichHB15} on the full data. Then, we encoded the source and the questions and the answers using these learned subword units. We reversed the source sequences before feeding them to the encoder in order to diminish the effect of vanishing gradients. We also applied padding and truncation to accommodate for bucketing.
%\footnote{Bucketing is a standard technique which is used to speed up training.}
%See the TensorFlow tutorial on sequence to sequence}. 
We then trained the seq2seq model using stochastic gradient descent.
%algorithm is then trained in the standard way by stochastic gradient descent which minimizes perplexity.

Every 2,000 iterations, we evaluated the current model with the metrics from Section~\ref{sec:modelselection}. These metrics are later used to select the model that is most suitable for our task, thus avoiding overfitting on the training data.

In our experiments, we used the following general parameter settings:
(\emph{i})~vocabulary size: 40,000 subword units; (\emph{ii})~dimensionality of the embedding vectors: 512; (\emph{iii})~RNN cell: 2-layered GRU cell with 512 units; (\emph{iv})~minibatch size: 80; (\emph{v})~learning rate: 0.5; (\emph{vi})~buckets: [(5, 10), (10, 15), (20, 25), (40,45)]. 
%\begin{itemize}
%\item vocabulary size - 40000 subword units
%\item size of embeddings vectors - 256 for the smaller, 512 for the larger 
%\item RNN cell - 2-layered GRU cell with 256 units for the smaller and 512 for the larger model
%\item batch size - 80
%\item learning rate - 0.5 
%\item buckets - [(5, 10), (10, 15), (20, 25), (40,45)]
%\end{itemize}

\subsection{Results and Discussion}

% \subsection{Comparing seq2seq to seq2seq with model selection \alex{Martin}}
% 1) best stopping criteria are rows 4,6 (which is expected). The fact that row 6 is best   is confirmed on DEV, too; 2) Best technique for computing scores selected on DEV (column G) has second best MAP on test (column AA), so it is consistent; 3) let's compare row 6 and BLUE (row 15) as stopping criteria according to BLUE score (columns AJ, AK): BLUE score in row 6 is better than the one in row 15; if we compare row 6 with BLUE_ALL (row 16), the latter has higher BLUE but it is using way more data (correct?)
% Several methods:
% \begin{itemize}
% \item seq2seq ranking model using (i) cosine similarity or  (2) BLEU score.
% \item Select the model based on accuracy, MAP, F1.\\
% \alex{Here the problem is that we do not have a threshold to apply to compute F1 and accuracy. So I guess we can only use MAP. An alternative idea is ROC}
% \item we can also do something creative like summation of the BLUE (or cosine) scores for positive minus the ones for negative.
% \item evaluate the system again probably only MAP/ROC as for the other measures we need a threshold that we do not have.
% \end{itemize}

In our first experiment, we explore the performance of seq2seq models produced by optimizing MAP using the different variants of the similarity function $r()$ from Section~\ref{sec:modelselection}, with MAP for model selection. 
%First of all, we note that all of the strategies beat the baseline-tfidf strategy, which shows that the seq2seq model is capable of 
The results in Table~\ref{tab:scoringfunctions} show that \textbf{TF$\times$IDF+qc-sim} performs best.
The results are consistent on the development (63.56) and on the test datasets (63.45).
The absolute improvement with respect to \textbf{cos-embeddings+qc-sim} is +0.59 on the development and +0.55 on the test dataset, respectively. 
% similarity function between utterance and comment/question, both on Development and Test sets. 

\begin{table}[t]
  \centering
  \begin{tabular}{|l|cc|}\hline
    & \multicolumn{2}{|c|}{\bf MAP} \\
  \bf Ranking Metric  & \bf Dev   & \bf Test \\\hline
  TF$\times$IDF+qc-sim    & 63.56 & 63.45 \\
  TF$\times$IDF   & 62.46 & 62.03 \\
  \hline
  cos-embeddings+qc-sim & 62.97 & 62.90 \\
  cos-embeddings  & 62.21   & 62.13 \\
  \hline
  bm25+qc-sim   & 62.81 &   61.96 \\
  bm25      & 62.88   & 61.77 \\
  \hline
  BLEU+qc-sim   & 62.67 &   62.73 \\
  BLEU      & 59.94 &   59.82 \\
  \hline
  Avg     & 62.84 &   62.33 \\
% baseline-tfidf &  59.50 & - \\
%   meanAvgBLEU 0.026898258 0.0283040228
%   BLEU_POS  0.0876761886  0.0816473398
%   BLEU_ALL  0.1743515877  0.1706251646
      \hline
  \end{tabular}
  \caption{MAP score for the ranking strategies defined in Section~\ref{sec:modelselection}, evaluated on the development and on the test datasets. \label{tab:scoringfunctions}}
\end{table}
      
In a second experiment, we compared model selection strategies when optimizing for MAP (\textbf{TF$\times$IDF+qc-sim}) vs. BLEU vs. seq2seq loss. We further report the results for a baseline for the SemEval2016 Task 3, subtask A~\cite{nakov-EtAl:2016:SemEval2}, which picks a random order for the answers in the target question-answer thread. 
The results are shown on Table~\ref{tab:results}. 
For each model selection criterion, we report its performance and statistics on the test dataset about the model that was best-performing on the development dataset. 

\noindent We can see that doing model selection according to MAP yielded not only the highest ranking performance of $63.45$ but also the best BLEU score. 
This is even more striking if we consider that BLEU tends to favor longer answers, but the average length of the seq2seq answers is $10.56$ for MAP and $16.31$ for BLEU score. 
Thus, we have shown that optimizing for an extrinsic evaluation measure that evaluates how good we are at telling \emph{Good} from \emph{Bad} answers works better than optimizing for BLEU. %(which has been traditionally used to evaluate dialog systems).

\begin{table*}[tbh]
  \centering
  \normalsize
  \begin{tabular}{|c|l|cccc|r|}
      \hline
    & &   \multicolumn{4}{|c|}{\bf \# \textit{Good} answers according to} &\\
  & \bf Optimizing for & \bf Ann. 1 & \bf Ann. 2  & \bf Ann. 3& \bf Ann. 4  & \multicolumn{1}{|c|}{\bf Avg.} \\\hline
   \multicolumn{7}{|c|}{\it questions 1-50} \\
  \hline  
1 & MAP & 23  & 29 &  21  & 26  & 24.75 (49.50\%) \\
2 & BLEU & 8  & 15  & 11  & 8 & 10.50 (21.00\%) \\
3 & seq2seq loss & 18 & 26  & 21  & 28  & 23.25 (46.50\%)\\
\hline
   \multicolumn{7}{|c|}{\it questions 51-100} \\
  \hline  
4 & MAP       & 28  & 20  & 13  & 29  & 22.50 (45.00\%)\\
5 & BLEU      &   9 &   5 &   2 &   6   & 5.50 (11.00\%) \\
6 & seq2seq loss  &   25  &   14  &   11  & 24  & 18.50 (37.00\%)\\
\hline
   \multicolumn{7}{|c|}{\it questions 1-100} \\
  \hline  
7 & MAP       & 51  & 49    & 34  & 55 &  47.25 (47.25\%) \\
8 & BLEU      & 17  & 20    & 13  & 14 &  16.00 (16.00\%)  \\
9 & seq2seq loss  & 43  & 40    & 32  & 52 &  41.75 (41.75\%) \\ 
\hline
\end{tabular}
  \caption{Number of good answers according to manual annotation of the answers to 50+50 questions by the three models from Table~\ref{tab:results}. \label{tab:results50}}
\end{table*}

\begin{table}[h]
  \centering
  \small
  \begin{tabular}{lll}
      \bf \# & \bf Model & \bf Example\\

            \hline
      1 & & \it How good are Karwa services?\\
      & BLEU & good luck\\
      & seq2seq & good luck\\
      & MAP & \bf karwa is good .\\

      \hline
      2 & & \it Which is the best spa in Doha?\\
      & BLEU & lulu\\
      & seq2seq & \bf al sharq spa\\
      & MAP & \bf al sharq spa\\

      \hline
      3 & & \it Which is better: Qtel or Vodafone?\\
      & BLEU & go for the bank\\
      & seq2seq & \bf vodafone\\
      & MAP & vodafone is better than vodafone .\\

      \hline
      4 & & \it Can you recommend a good Vet in Doha?\\
      & BLEU & good luck\\
      & seq2seq & \bf dr . choc\\
      & MAP & \bf dr . choc\\

      \hline
      5 & & \it Where can I go running; other than \\
      & & \it the Corniche?\\
      & BLEU & i think it is a good place to go to doha .\\
      & seq2seq & corniche\\
      & MAP & corniche\\

      \hline
      6 & & \it Are there any cameras on the \\
      & & \it roundabouts in Qatar?\\
      & BLEU & i think it is a good offer for a few months .\\
      & seq2seq & there are many cars in qatar .\\
      & MAP & \bf there is no radar in qatar .\\

      \hline
      7 & & \it What is the best mall in Doha to buy \\
      & & \it good furniture?\\
      & BLEU & lulu is good\\
      & seq2seq & carrefour\\
      & MAP & \bf ikea\\
      \hline
\end{tabular}
  \caption{Output of our chatbot when using different model selection strategies for some simplified questions from Qatar Living, which we used for the first part of our manual evaluation. The good answers are shown in bold. \label{tab:manual:examples}}
\end{table}

\begin{table}[h]
  \centering
  \small
  \begin{tabular}{lll}
      \bf \# & \bf Model & \bf Example\\

      \hline
      8 & & \it Do you like fast food?\\
      & BLEU & \bf i like the one in doha\\
      & seq2seq & \bf i do n ' t eat food .\\
      & MAP & \bf i hate food .\\

      \hline
      9 & & \it What is your favorite beach in Qatar?\\
      & BLEU & i think it is a good place\\
      & seq2seq & \bf sealine beach\\
      & MAP & \bf sealine beach\\

      \hline
      10 & & \it Tell me bot, what do you know \\
      & & \it about Qatar?\\
      & BLEU & i think it is a good offer to qatar .\\
      & seq2seq & qatar living\\
      & MAP & \bf qatar is a country .\\

      \hline
      11 & & \it What are the things about Qatar you\\
         & & \it like the most?\\
      & BLEU & i think it ' s a good one ...\\
      & seq2seq & qatar is a country .\\
      & MAP & \bf qatar is the best place to live .\\

      \hline
      12 & & \it Are you looking forward to the FIFA\\ 
         & & \it World Cup in 2022?\\
      & BLEU & i think it was a good one in doha\\
      & seq2seq & i will win the world cup .\\
      & MAP & i will win the world cup .\\

      \hline
      13 & & \it Do you think the fast food in Qatar is of \\
      & & \it good quality?\\
      & BLEU & i think it is a good place to go to doha .\\
      & seq2seq & \bf food is good .\\
      & MAP & food is good for food .\\

      \hline
      14 & & \it What is the most useful thing that you \\
      & & \it can tell me about life in Qatar?\\
      & BLEU & i love my life ...\\
      & seq2seq & life is life ... : - rrb -\\
      & MAP & \bf life is beautiful ...\\

      \hline
      15 & & \it Do you need to Google to give me \\
      & & \it an answer?\\
      & BLEU & i think you can get a good one in \\
      & & your home\\
      & seq2seq & i will give you the answer .\\
      & MAP & \bf i would like to answer .\\

	\hline

  \end{tabular}
  \caption{Output of our chatbot 
%  when using different model selection strategies 
for some new, conversational questions related to Qatar, which we created and used for the second part of our manual evaluation. The good answers are shown in bold.
  \label{tab:manual:conversational-examples}}
\end{table}

\section{Manual Evaluation and Error Analysis} \label{sec:erroranalysis} 

We evaluated the three approaches in Table~\ref{tab:results} on 100 relatively short questions.\footnote{The questions and the outputs of the different models are available at \url{http://goo.gl/w9MZfv}} 
First, we randomly selected 50 questions from the test set of SemEval-2016 Task 3. However, we did not use the original questions, as they can contain multiple sentences and thus can be too long; instead, we selected a single sentence that contains the core of the question (and in some cases, we simplified it a bit further). We further created 50 new questions, which are more personal and conversational in nature, but are still generally related to Qatar.
The answers produced by the three systems for these 100 questions were evaluated independently by four annotators, who judged whether each of the answers is good.

\subsection{Quantitive Analysis}

Table~\ref{tab:results50} reports the number of good answers that each of the annotators has judged to be good when the model is selected based on BLEU, MAP, and seq2seq loss. The average of the four annotators suggests that optimizing for MAP yields the best overall results. 
Note that all systems perform slightly worse on the second set of questions. This should be expected as the latter are different from those used for training the models. 
Overall, the MAP-based system appears to be more robust, with only 2.25 points absolute decrease in performance (compared to 5 and 4.75 for the systems using BLEU and seq2seq loss, respectively).

%\begin{table*}[ht]
%  \centering
%  \begin{tabular}{|c|l|ccc|c|}
%      \hline
%   & &   \multicolumn{3}{|c|}{\# \textit{Good} answers according to} &\\
%  & \bf Optimizing for & \bf Ann. 1  & \bf Ann. 2  & \bf Ann. 3  & \bf Avg. \\
%  \hline  
%1 & MAP      & 24  & 26    & 28  &   26.0  \\
%2 & BLEU     & 12  & 1   & 8 &   7.0 \\
%3 & seq2seq loss & 19  & 28    & 24  & 23.7  \\ \hline
%\end{tabular}
%  \caption{Number of \textit{Good} answers according to manual annotation of the answers to 50 questions by the three seq2seq systems in Table~\ref{tab:results}. \label{tab:results50}}
%\end{table*}

%\subsection{Qualitative Analysis} 
%In this section we analyze the quality and the shortcomings of the generated answers in the manual evaluation. Table \ref{tab:manual:examples} and Table \ref{tab:manual:conversational-examples} list a series of interesting cases on which we will comment. \\

\subsection{Qualitative Analysis}

We now analyze the quality of the generated answers from the manual evaluation. Instead of looking at overall numbers, here we look at some interesting cases, shown in Tables \ref{tab:manual:examples} and \ref{tab:manual:conversational-examples}. 

First, we can confirm that the answers generated by the model that was optimized for BLEU seem to be the worst. We attribute this to the relatively early iteration when the optimal BLEU occurs and also to the nature of the BLEU metric. BLEU tries to optimize for $n$-gram matching. Thus, the selected model ultimately prefers longer utterances, while the other two models focus on providing a short focused answer; this is especially true for the first part of the manual test set as shown in Table~\ref{tab:manual:examples},
where we can find ``safe'' answers with stopwords, which do not have much informative content, but are a good bet, e.g., ``good luck'', ``I think that'', ``it is good to'', etc. 

In examples 1 and 6, we can see that only the MAP-based model addressed the question directly. The other models are a better fit to the language model, and thus failed to produce the target named entity. In example 1, they produced a generic answer, which can be given in response to many questions.

\noindent Example 5 shows how the models have trouble handling exclusion/negation. The model was able to copy the named entity, which is generally a good thing to do, but not here. If the question was simply ``Where can I go running?'', the answer would have been good. 

Note that the responses to the questions from the second group are more personal, e.g., they start with ``I think'', ``I will'', etc.
Finally, we can see the well-known problem with seq2seq models: repetition. This is evident in examples 3, 13, 14. 

%In \cite{DBLP:journals/corr/ShaoGBGSK17} they try to tackle this problem by introducing a \emph{glimpse model} which pays attention to the previous output.

%We have created 50 new questions 
%\begin{itemize}
%\item Evaluate 10 questions: Basic seq2seq vs seq2seqMS (best model selection)
%\item elaborate on some interesting case, where seq2seq is doing well and the other wrong.
%\end{itemize}

\section{Conclusion} \label{sec:conclusion} 

Building dialog systems, e.g., in the form of chatbots, has attracted a lot of attention recently and thus has become a crucial investment for many companies. Progress in neural networks, especially in seq2seq models, has made it possible to quickly and directly learn chatbots from data. However, the availability of domain-specific training data coming from real dialogs is a critical bottleneck for the development of dialog systems. 

We addressed the problem by producing training data from Community Question Answering (CQA) forums. We further applied sentence selection based on word embeddings in order to retain only meaningful pairs of short texts, which can be more effectively used for training seq2seq models. 

Moreover, we introduced the use of extrinsic evaluation based on a CQA task using MAP to select the most effective models among those generated during training, using a development set of good user answers available from the cQA data. 

\noindent We also used MAP to perform automatic evaluation of system accuracy against the test set annotated for the CQA task. This was not explored before.

Finally, we carried out manual evaluation with four different annotators on two different sets of questions: the first set used simplified questions taken from the CQA data, whereas the second one was composed of new, conversational-style questions that we generated. Thus, the questions of the second set are rather different from those used to train the systems; yet, they are about topics that are generally discussed in the training data.

We have found that the seq2seq model can learn from CQA-derived data, producing accurate answers when answering forum questions according to automatic and manual evaluation, with MAP of 63.45, and accuracy of 49.50, respectively. Moreover, the accuracy on completely new questions drops by only few points, i.e., to 47.25,~according to our manual evaluation.

\noindent Interestingly, our model selection is more accurate than using the loss of the seq2seq model, and performs much better than BLEU. Indeed, the latter seems not to be very appropriate for evaluating chatbots,
% (at least such trained on CQA data), 
as our manual analysis shows.

In future work, we would like to study new methods for selecting data, so that the overall system accuracy can improve further. 
We also plan to try sub-word embedding representations \cite{DBLP:journals/corr/BojanowskiGJM16} that could better capture typos, which are common in Web forums,
and to experiment with other languages such as Arabic.
Last but not least, we want to explore adversarial dialog training and evaluation \cite{bruni-fernandez:2017:W17-55,adversarial:dialog:2016,li-EtAl:2017:EMNLP20175,SeqGan:2017}.

\section*{Acknowledgments}
This research was performed by the Arabic Language Technologies group at Qatar Computing Research Institute, HBKU\@, within the Interactive sYstems for Answer Search project ({\sc Iyas}).

\bibliography{sigproc}

\begin{thebibliography}{}
\expandafter\ifx\csname natexlab\endcsname\relax\def\natexlab#1{#1}\fi

\bibitem[{Bahdanau et~al.(2015)Bahdanau, Cho, and
  Bengio}]{DBLP:journals/corr/BahdanauCB14}
Dzmitry Bahdanau, Kyunghyun Cho, and Yoshua Bengio. 2015.
\newblock Neural machine translation by jointly learning to align and
  translate.
\newblock In {\em Proceedings of 3rd International Conference on Learning
  Representations\/}. San Diego, California, USA, ICLR~'15.

\bibitem[{Bojanowski et~al.(2017)Bojanowski, Grave, Joulin, and
  Mikolov}]{DBLP:journals/corr/BojanowskiGJM16}
Piotr Bojanowski, Edouard Grave, Armand Joulin, and Tomas Mikolov. 2017.
\newblock Enriching word vectors with subword information.
\newblock {\em Transactions of the Association for Computational Linguistics\/}
  5:135--146.

\bibitem[{Bruni and Fernandez(2017)}]{bruni-fernandez:2017:W17-55}
Elia Bruni and Raquel Fernandez. 2017.
\newblock Adversarial evaluation for open-domain dialogue generation.
\newblock In {\em Proceedings of the 18th Annual SIGdial Meeting on Discourse
  and Dialogue\/}. Saarbruecken, Germany, SIGDIAL~'17, pages 284--288.

\bibitem[{Cho et~al.(2014)Cho, van Merrienboer, Gulcehre, Bahdanau, Bougares,
  Schwenk, and Bengio}]{cho-EtAl:2014:EMNLP2014}
Kyunghyun Cho, Bart van Merrienboer, Caglar Gulcehre, Dzmitry Bahdanau, Fethi
  Bougares, Holger Schwenk, and Yoshua Bengio. 2014.
\newblock Learning phrase representations using {RNN} encoder--decoder for
  statistical machine translation.
\newblock In {\em Proceedings of the Conference on Empirical Methods in Natural
  Language Processing\/}. Doha, Qatar, EMNLP~'14, pages 1724--1734.

\bibitem[{Gu et~al.(2016)Gu, Lu, Li, and Li}]{gu-EtAl:2016:P16-1}
Jiatao Gu, Zhengdong Lu, Hang Li, and Victor~O.K. Li. 2016.
\newblock Incorporating copying mechanism in sequence-to-sequence learning.
\newblock In {\em Proceedings of the 54th Annual Meeting of the Association for
  Computational Linguistics\/}. Berlin, Germany, ACL~'16, pages 1631--1640.

\bibitem[{Guzm\'{a}n et~al.(2016)Guzm\'{a}n, M\`{a}rquez, and
  Nakov}]{guzman-marquez-nakov:2016:P16-2}
Francisco Guzm\'{a}n, Llu\'{i}s M\`{a}rquez, and Preslav Nakov. 2016.
\newblock Machine translation evaluation meets community question answering.
\newblock In {\em Proceedings of the 54th Annual Meeting of the Association for
  Computational Linguistic\/}. Berlin, Germany, ACL~'16, pages 460--466.

\bibitem[{Hochreiter and Schmidhuber(1997)}]{Hochreiter:1997:LSTM}
Sepp Hochreiter and J\"{u}rgen Schmidhuber. 1997.
\newblock Long short-term memory.
\newblock {\em Neural Comput.\/} 9(8):1735--1780.

\bibitem[{Hoque et~al.(2016)Hoque, Joty, M\`{a}rquez, Barr\'{o}n-Cede\~{n}o,
  Da~San~Martino, Moschitti, Nakov, Romeo, and
  Carenini}]{hoque-EtAl:2016:COLINGDEMO}
Enamul Hoque, Shafiq Joty, Llu\'{i}s M\`{a}rquez, Alberto
  Barr\'{o}n-Cede\~{n}o, Giovanni Da~San~Martino, Alessandro Moschitti, Preslav
  Nakov, Salvatore Romeo, and Giuseppe Carenini. 2016.
\newblock An interactive system for exploring community question answering
  forums.
\newblock In {\em Proceedings of the 26th International Conference on
  Computational Linguistics\/}. Osaka, Japan, COLING~'16, pages 1--5.

\bibitem[{Kannan and Vinyals(2016)}]{adversarial:dialog:2016}
Anjuli Kannan and Oriol Vinyals. 2016.
\newblock {SeqGAN}: Sequence generative adversarial nets with policy gradient.
\newblock In {\em Proceedings of the NIPS 2016 Workshop on Adversarial
  Training\/}. Barcelona, Spain.

\bibitem[{Kim(2014)}]{kim:2014:EMNLP2014}
Yoon Kim. 2014.
\newblock Convolutional neural networks for sentence classification.
\newblock In {\em Proceedings of the Conference on Empirical Methods in Natural
  Language Processing\/}. Doha, Qatar, EMNLP~'14, pages 1746--1751.

\bibitem[{Lavie and Agarwal(2007)}]{Lavie:2007:MAM:1626355.1626389}
Alon Lavie and Abhaya Agarwal. 2007.
\newblock Meteor: An automatic metric for {MT} evaluation with high levels of
  correlation with human judgments.
\newblock In {\em Proceedings of the Second Workshop on Statistical Machine
  Translation\/}. Prague, Czech Republic, WMT~'07, pages 228--231.

\bibitem[{Li et~al.(2016)Li, Galley, Brockett, Gao, and
  Dolan}]{li-EtAl:2016:N16-11}
Jiwei Li, Michel Galley, Chris Brockett, Jianfeng Gao, and Bill Dolan. 2016.
\newblock A diversity-promoting objective function for neural conversation
  models.
\newblock In {\em Proceedings of the North American Chapter of the Association
  for Computational Linguistics: Human Language Technologies\/}. San Diego,
  California, USA, NAACL-HLT~'16, pages 110--119.

\bibitem[{Li et~al.(2017)Li, Monroe, Shi, Jean, Ritter, and
  Jurafsky}]{li-EtAl:2017:EMNLP20175}
Jiwei Li, Will Monroe, Tianlin Shi, S\'ebastien Jean, Alan Ritter, and Dan
  Jurafsky. 2017.
\newblock Adversarial learning for neural dialogue generation.
\newblock In {\em Proceedings of the 2017 Conference on Empirical Methods in
  Natural Language Processing\/}. Copenhagen, Denmark, EMNLP~'17, pages
  2147--2159.

\bibitem[{Liu et~al.(2016)Liu, Lowe, Serban, Noseworthy, Charlin, and
  Pineau}]{liu-EtAl:2016:EMNLP20163}
Chia-Wei Liu, Ryan Lowe, Iulian Serban, Mike Noseworthy, Laurent Charlin, and
  Joelle Pineau. 2016.
\newblock How {NOT} to evaluate your dialogue system: An empirical study of
  unsupervised evaluation metrics for dialogue response generation.
\newblock In {\em Proceedings of the Conference on Empirical Methods in Natural
  Language Processing\/}. Austin, Texas, USA, EMNLP~'16, pages 2122--2132.

\bibitem[{Lowe et~al.(2015)Lowe, Pow, Serban, and
  Pineau}]{DBLP:conf/sigdial/LowePSP15}
Ryan Lowe, Nissan Pow, Iulian Serban, and Joelle Pineau. 2015.
\newblock The {U}buntu dialogue corpus: A large dataset for research in
  unstructured multi-turn dialogue systems.
\newblock In {\em Proceedings of the 16th Annual Meeting of the Special
  Interest Group on Discourse and Dialogue\/}. Prague, Czech Republic,
  SIGDIAL~'15, pages 285--294.

\bibitem[{Mihaylov et~al.(2017)Mihaylov, Balchev, Kiprov, Koychev, and
  Nakov}]{Mihaylov:2017:LGP:3077136.3080757}
Todor Mihaylov, Daniel Balchev, Yasen Kiprov, Ivan Koychev, and Preslav Nakov.
  2017.
\newblock Large-scale goodness polarity lexicons for community question
  answering.
\newblock In {\em Proceedings of the 40th International ACM SIGIR Conference on
  Research and Development in Information Retrieval\/}. Tokyo, Japan,
  SIGIR~'17, pages 1185--1188.

\bibitem[{Mihaylov and Nakov(2016)}]{mihaylov-nakov:2016:SemEval}
Todor Mihaylov and Preslav Nakov. 2016.
\newblock {SemanticZ} at {SemEval}-2016 task 3: Ranking relevant answers in
  community question answering using semantic similarity based on fine-tuned
  word embeddings.
\newblock In {\em Proceedings of the 10th International Workshop on Semantic
  Evaluation\/}. San Diego, California, USA, SemEval~'16, pages 804 -- 811.

\bibitem[{Mihaylova et~al.(2016)Mihaylova, Gencheva, Boyanov, Yovcheva,
  Mihaylov, Hardalov, Kiprov, Balchev, Koychev, Nakov, Nikolova, and
  Angelova}]{mihaylova-EtAl:2016:SemEval}
Tsvetomila Mihaylova, Pepa Gencheva, Martin Boyanov, Ivana Yovcheva, Todor
  Mihaylov, Momchil Hardalov, Yasen Kiprov, Daniel Balchev, Ivan Koychev,
  Preslav Nakov, Ivelina Nikolova, and Galia Angelova. 2016.
\newblock {SUper} team at {SemEval}-2016 task 3: Building a feature-rich system
  for community question answering.
\newblock In {\em Proceedings of the 10th International Workshop on Semantic
  Evaluation\/}. San Diego, California, USA, SemEval~'16, pages 836--843.

\bibitem[{Mikolov et~al.(2013{\natexlab{a}})Mikolov, Sutskever, Chen, Corrado,
  and Dean}]{Mikolov:2013:DRW:2999792.2999959}
Tomas Mikolov, Ilya Sutskever, Kai Chen, Greg Corrado, and Jeffrey Dean.
  2013{\natexlab{a}}.
\newblock Distributed representations of words and phrases and their
  compositionality.
\newblock In {\em Proceedings of the 26th International Conference on Neural
  Information Processing Systems\/}. Lake Tahoe, Nevada, USA, NIPS'13, pages
  3111--3119.

\bibitem[{Mikolov et~al.(2013{\natexlab{b}})Mikolov, Yih, and
  Zweig}]{mikolov-yih-zweig:2013:NAACL-HLT}
Tomas Mikolov, Wen-tau Yih, and Geoffrey Zweig. 2013{\natexlab{b}}.
\newblock Linguistic regularities in continuous space word representations.
\newblock In {\em Proceedings of the 2013 Conference of the North American
  Chapter of the Association for Computational Linguistics: Human Language
  Technologies\/}. Atlanta, Georgia, USA, NAACL-HLT~'13, pages 746--751.

\bibitem[{Nakov et~al.(2017)Nakov, Hoogeveen, M\`{a}rquez, Moschitti, Mubarak,
  Baldwin, and Verspoor}]{SemEval-2017:task3}
Preslav Nakov, Doris Hoogeveen, Llu\'{i}s M\`{a}rquez, Alessandro Moschitti,
  Hamdy Mubarak, Timothy Baldwin, and Karin Verspoor. 2017.
\newblock {SemEval}-2017 task 3: Community question answering.
\newblock In {\em Proceedings of the 11th International Workshop on Semantic
  Evaluation\/}. Vancouver, British Columbia, Canada, SemEval~'17, pages
  27--48.

\bibitem[{Nakov et~al.(2016{\natexlab{a}})Nakov, M\`{a}rquez, and
  Guzm\'{a}n}]{nakov-marquez-guzman:2016:EMNLP2016}
Preslav Nakov, Llu\'{i}s M\`{a}rquez, and Francisco Guzm\'{a}n.
  2016{\natexlab{a}}.
\newblock It takes three to tango: Triangulation approach to answer ranking in
  community question answering.
\newblock In {\em Proceedings of the Conference on Empirical Methods in Natural
  Language Processing\/}. Austin, Texas, USA, EMNLP~'16, pages 1586--1597.

\bibitem[{Nakov et~al.(2015)Nakov, M\`{a}rquez, Magdy, Moschitti, Glass, and
  Randeree}]{nakov-EtAl:2015:SemEval}
Preslav Nakov, Llu\'{i}s M\`{a}rquez, Walid Magdy, Alessandro Moschitti, Jim
  Glass, and Bilal Randeree. 2015.
\newblock {SemEval}-2015 task 3: Answer selection in community question
  answering.
\newblock In {\em Proceedings of the 9th International Workshop on Semantic
  Evaluation\/}. Denver, Colorado, USA, SemEval~'15, pages 269--281.

\bibitem[{Nakov et~al.(2016{\natexlab{b}})Nakov, M\`{a}rquez, Moschitti, Magdy,
  Mubarak, Freihat, Glass, and Randeree}]{nakov-EtAl:2016:SemEval2}
Preslav Nakov, Llu\'{i}s M\`{a}rquez, Alessandro Moschitti, Walid Magdy, Hamdy
  Mubarak, abed~Alhakim Freihat, Jim Glass, and Bilal Randeree.
  2016{\natexlab{b}}.
\newblock {SemEval}-2016 task 3: Community question answering.
\newblock In {\em Proceedings of the 10th International Workshop on Semantic
  Evaluation\/}. San Diego, California, USA, SemEval~'16, pages 525--545.

\bibitem[{Papineni et~al.(2002)Papineni, Roukos, Ward, and
  Zhu}]{Papineni:2002:BMA:1073083.1073135}
Kishore Papineni, Salim Roukos, Todd Ward, and Wei-Jing Zhu. 2002.
\newblock {BLEU}: A method for automatic evaluation of machine translation.
\newblock In {\em Proceedings of the 40th Annual Meeting on Association for
  Computational Linguistics\/}. Philadelphia, Pennsylvania, USA, ACL '02, pages
  311--318.

\bibitem[{Robertson and Zaragoza(2009)}]{Robertson:2009:PRF:1704809.1704810}
Stephen Robertson and Hugo Zaragoza. 2009.
\newblock The probabilistic relevance framework: {BM25} and beyond.
\newblock {\em Found. Trends Inf. Retr.\/} 3(4):333--389.

\bibitem[{See et~al.(2017)See, Liu, and Manning}]{summarization}
Abigail See, Peter~J. Liu, and Christopher~D. Manning. 2017.
\newblock Get to the point: Summarization with pointer-generator networks.
\newblock In {\em Proceedings of the 55th Annual Meeting of the Association for
  Computational Linguistics\/}. Vancouver, British Columbia, Canada, ACL~'17,
  pages 1073--1083.

\bibitem[{Sennrich et~al.(2016)Sennrich, Haddow, and
  Birch}]{DBLP:journals/corr/SennrichHB15}
Rico Sennrich, Barry Haddow, and Alexandra Birch. 2016.
\newblock Neural machine translation of rare words with subword units.
\newblock In {\em Proceedings of the 54th Annual Meeting of the Association for
  Computational Linguistics\/}. Berlin, Germany, ACL~'16, pages 1715--1725.

\bibitem[{Serban et~al.(2015)Serban, Lowe, Henderson, Charlin, and
  Pineau}]{DBLP:journals/corr/SerbanLCP15}
Iulian~Vlad Serban, Ryan Lowe, Peter Henderson, Laurent Charlin, and Joelle
  Pineau. 2015.
\newblock A survey of available corpora for building data-driven dialogue
  systems.
\newblock {\em CoRR\/} abs/1512.05742.

\bibitem[{Shang et~al.(2015)Shang, Lu, and Li}]{shang-lu-li:2015:ACL-IJCNLP}
Lifeng Shang, Zhengdong Lu, and Hang Li. 2015.
\newblock Neural responding machine for short-text conversation.
\newblock In {\em Proceedings of the 53rd Annual Meeting of the Association for
  Computational Linguistics and the 7th International Joint Conference on
  Natural Language Processing\/}. Beijing, China, ACL-IJCNLP~'15, pages
  1577--1586.

\bibitem[{Sordoni et~al.(2015)Sordoni, Galley, Auli, Brockett, Ji, Mitchell,
  Nie, Gao, and Dolan}]{sordoni-EtAl:2015:NAACL-HLT}
Alessandro Sordoni, Michel Galley, Michael Auli, Chris Brockett, Yangfeng Ji,
  Margaret Mitchell, Jian-Yun Nie, Jianfeng Gao, and Bill Dolan. 2015.
\newblock A neural network approach to context-sensitive generation of
  conversational responses.
\newblock In {\em Proceedings of the Conference of the North American Chapter
  of the Association for Computational Linguistics: Human Language
  Technologies\/}. Denver, Colorado, USA, NAACL-HLT~'15, pages 196--205.

\bibitem[{Sutskever et~al.(2014)Sutskever, Vinyals, and
  Le}]{sutskever2014sequence}
Ilya Sutskever, Oriol Vinyals, and Quoc~V. Le. 2014.
\newblock Sequence to sequence learning with neural networks.
\newblock In {\em Proceedings of the 27th International Conference on Neural
  Information Processing Systems\/}. Montr{\'e}al, Qu{\'e}bec, Canada,
  NIPS~'14, pages 3104--3112.

\bibitem[{Vinyals and Le(2015)}]{DBLP:journals/corr/VinyalsL15}
Oriol Vinyals and Quoc~V. Le. 2015.
\newblock A neural conversational model.
\newblock {\em CoRR\/} abs/1506.05869.

\bibitem[{Vinyals et~al.(2017)Vinyals, Toshev, Bengio, and
  Erhan}]{Vinyals:2017:STL:3069214.3069249}
Oriol Vinyals, Alexander Toshev, Samy Bengio, and Dumitru Erhan. 2017.
\newblock Show and tell: Lessons learned from the 2015 {MSCOCO} image
  captioning challenge.
\newblock {\em IEEE Trans. Pattern Anal. Mach. Intell.\/} 39(4):652--663.

\bibitem[{Yu et~al.(2017)Yu, Zhang, Wang, and Yu}]{SeqGan:2017}
Lantao Yu, Weinan Zhang, Jun Wang, and Yong Yu. 2017.
\newblock {SeqGAN}: Sequence generative adversarial nets with policy gradient.
\newblock In {\em Proceedings of the 31st Conference on Artificial
  Intelligence\/}. San Francisco, California, USA, AAAI~'17, pages 2852--2858.

\end{thebibliography}
\bibliographystyle{acl_natbib}

\appendix

\end{document}